\PassOptionsToPackage{table, svgnames}{xcolor}
\documentclass{article} 
\usepackage{iclr2025_delta,times}

\usepackage{amsmath,amsfonts,bm}

\def\eqref#1{equation~\ref{#1}}

\def\1{\bm{1}}

\def\vx{{\bm{x}}}

\DeclareMathAlphabet{\mathsfit}{\encodingdefault}{\sfdefault}{m}{sl}
\SetMathAlphabet{\mathsfit}{bold}{\encodingdefault}{\sfdefault}{bx}{n}

\usepackage{hyperref}
\usepackage{url}

\usepackage[utf8]{inputenc}
\usepackage{multirow}
\usepackage{tablefootnote}
\usepackage{makecell}
\usepackage{enumitem}
\usepackage{soul}
\usepackage[export]{adjustbox}
\usepackage{bbm}
\usepackage{bbding}
\usepackage[toc,page,header]{appendix}
\usepackage{minitoc}
\usepackage{graphicx}
\usepackage{booktabs} 
\usepackage{subfig}
\usepackage{cleveref}
\usepackage{tikz}
\usepackage{mathtools}
\usepackage{tcolorbox}

\newcommand*\circled[1]{\tikz[baseline=(char.base)]{\node[shape=circle,draw,inner sep=0.5pt] (char) {#1};}}

\newcommand{\titlecontent}{How Well Does Your Tabular Generator Learn the Structure of Tabular Data?}
\title{\titlecontent}

\author{Xiangjian Jiang$^{1}$, Nikola Simidjievski$^{1,2}$ \& Mateja Jamnik$^{1}$ \\
$^{1}$\normalfont{Department of Computer Science and Technology} \\
$^{2}$\normalfont{PBCI, Department of Oncology} \\
University of Cambridge\\
Cambridge, UK \\
\texttt{\{xj265,ns779,mj201\}@cam.ac.uk}
}

\iclrfinalcopy 
\begin{document}
\doparttoc
\faketableofcontents

\maketitle

\begin{abstract}
Heterogeneous tabular data poses unique challenges in generative modelling due to its fundamentally different underlying data structure compared to homogeneous modalities, such as images and text. Although previous research has sought to adapt the successes of generative modelling in homogeneous modalities to the tabular domain, defining an effective generator for tabular data remains an open problem. One major reason is that the evaluation criteria inherited from other modalities often fail to adequately assess whether tabular generative models effectively capture or utilise the unique structural information encoded in tabular data.
In this paper, we carefully examine the limitations of the prevailing evaluation framework and introduce \textbf{TabStruct}, a novel evaluation benchmark that positions structural fidelity as a core evaluation dimension. Specifically, TabStruct evaluates the alignment of causal structures in real and synthetic data, providing a direct measure of how effectively tabular generative models learn the structure of tabular data.
Through extensive experiments using generators from eight categories on seven datasets with expert-validated causal graphical structures, we show that structural fidelity offers a task-independent, domain-agnostic evaluation dimension.
Our findings highlight the importance of tabular data structure and offer practical guidance for developing more effective and robust tabular generative models.
Code is available at \url{https://github.com/SilenceX12138/TabStruct}.
\looseness-1
\end{abstract}

\section{Introduction}
Tabular data generation is a cornerstone of many real-world machine learning tasks~\citep{borisov2022deep, fang2024large}, ranging from training data augmentation~\citep{margeloiutabebm, cui2024tabular} to missing data imputation~\citep{zhangmixed}. These applications highlight the importance of building powerful models capable of generating high-quality synthetic tabular data, which necessitates an appropriate understanding of the underlying data structure. For instance, textual data conforms to the distributional hypothesis, and thus the autoregressive process can be a natural and effective approach for text generation~\citep{zhao2023survey, sahlgren2008distributional}. 
In contrast, tabular data poses unique challenges due to its heterogeneity -- the features within a dataset typically have varying types and semantics, with feature sets that can differ across datasets~\citep{grinsztajn2022tree, shi2024tabdiff}. 
Recent work in tabular foundation predictors demonstrates that (causal) structure can be an effective prior for tabular data structure~\citep{hollmann2025accurate}, which is fundamentally different to homogeneous modalities like text or images. As such, it is important to investigate how effectively existing tabular generative models capture and leverage the tabular data structure. 

Prior work~\citep{hansen2023reimagining, zhangmixed, margeloiutabebm} has proposed tabular generative models spanning multiple categories for high-quality synthetic data. However, a fair and comprehensive benchmarking framework remains absent. Specifically, existing benchmarks exhibit three primary limitations:
\textbf{(i)~Lack of evaluating the tabular data structure.}  
The mainstream benchmarks primarily adopt evaluation dimensions from homogeneous modalities, including density estimation~\citep{alaa2022faithful}, downstream utility~\citep{xu2019modeling}, and privacy preservation~\citep{kotelnikov2023tabddpm}. While these metrics have proven effective for other modalities, they fail to fully assess whether tabular generative models capture the unique structural information of tabular data.
\textbf{(ii)~Potentially biased evaluation.}  
Beyond overlooking structural information, certain conventional evaluation metrics may introduce bias (see \Cref{sec:conventional-eval-dim} for more details). For instance, evaluating synthetic data based on downstream utility depends heavily on the choice of the performance metric as well as the downstream models and tasks~\citep{hansen2023reimagining, margeloiutabebm}, which may obscure the true capabilities of tabular generative models.
\textbf{(iii)~Limited coverage of tabular generative models.}  
Existing benchmarks often evaluate a narrow range of tabular generative models, limiting their ability to provide a comprehensive comparison of model performance across the broader landscape of tabular generative modelling. \Cref{appendix:literature-review} further summarises the scope of the evaluation metrics and generators in TabStruct and existing benchmarks.
In this paper, we aim to address these gaps by developing a systematic and comprehensive evaluation framework for existing tabular generative models.

We introduce \textbf{TabStruct} (\Cref{fig:framework}), a novel benchmark framework designed to comprehensively evaluate tabular generative models across diverse metrics and model categories. TabStruct is characterised by three core concepts. 
Firstly, TabStruct positions structural fidelity as a core evaluation dimension, and quantifies it through the alignment of feature independence relationships between real and synthetic data.
Secondly, TabStruct retains the conventional evaluation metrics and investigates their interplay with structural fidelity.
Thirdly, TabStruct includes eight generator categories, ensuring holistic and robust benchmarking results.

\looseness-1
Our contributions can be summarised as follows: 
\textbf{\textcolor{ForestGreen}{\circled{1}} Conceptual}~(\Cref{sec:method}): We propose TabStruct, a novel benchmark framework that integrates structural fidelity as a core evaluation dimension for tabular generative models.
\textbf{\textcolor{ForestGreen}{\circled{2}} Empirical}~(\Cref{sec:exp}): We quantitatively analyse the model capabilities across four dimensions and provide actionable insights for designing more robust tabular generative models.
\textbf{\textcolor{ForestGreen}{\circled{3}} Technical}: We will release TabStruct, including the benchmark suite, the associated codebase, and all raw experimental results. This open-source library will enable researchers and practitioners to evaluate their models efficiently and comprehensively with a standardised framework.

\section{TabStruct Benchmark Framework}
\label{sec:method}

\begin{figure}[!t]
    \centering
    \includegraphics[clip, trim=2cm 0cm 2cm 0cm, width=\textwidth]{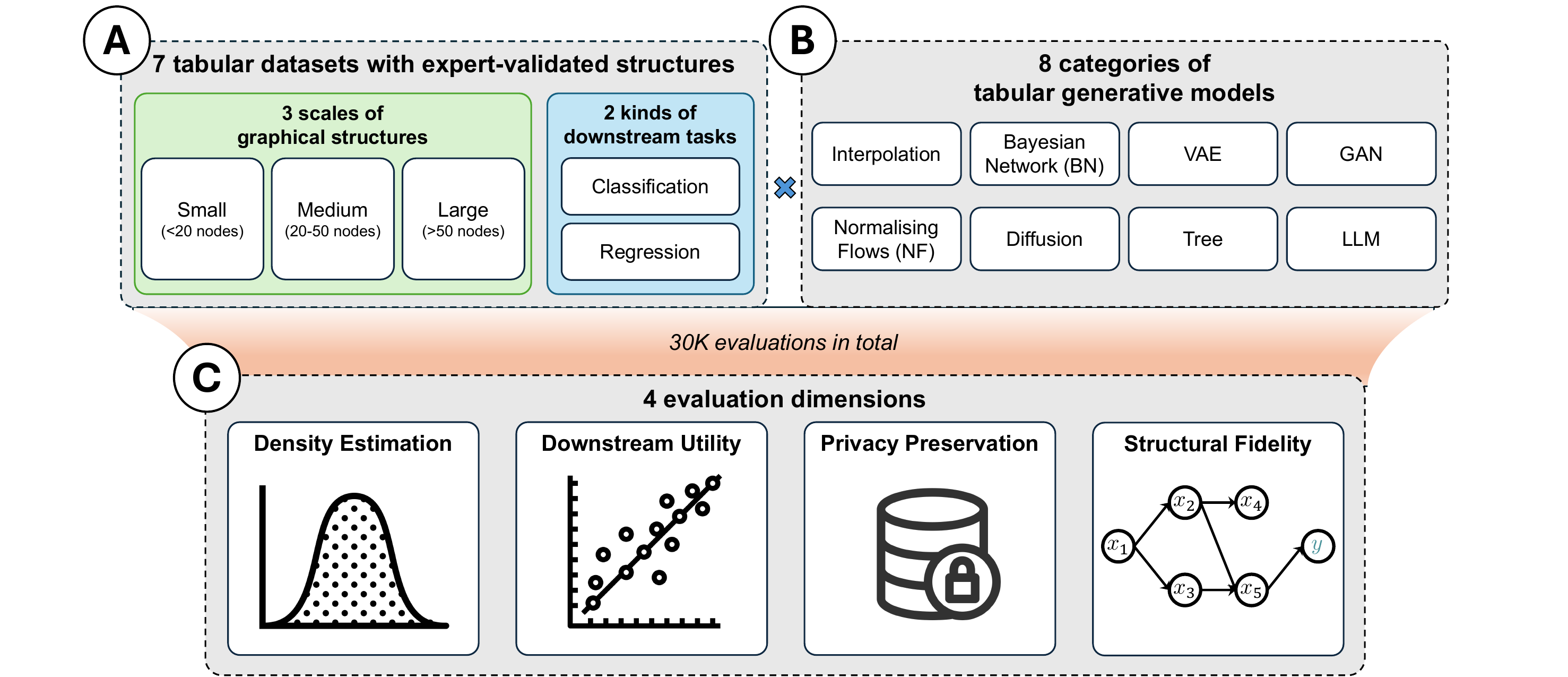}
    \caption{\textbf{The overview of the TabStruct evaluation framework.} \textbf{(A)}~Given the graphical structures (i.e., structural causal models) validated by domain experts, we perform prior sampling on these graphs to generate a full dataset $\mathcal{D}$. \textbf{(B)}~We train tabular generative models on the training split $\mathcal{D}_{\text{ref}} \subset \mathcal{D}$. We then  generate synthetic data $\mathcal{D}_{\text{syn}}$ with the fitted models. \textbf{(C)}~We evaluate the quality of synthetic data by comparing $\mathcal{D}_{\text{ref}}$ and $\mathcal{D}_{\text{syn}}$ across four dimensions.}
\label{fig:framework}
\end{figure}

\Cref{fig:framework} provides an overview of the TabStruct framework. We first describe our problem setup (\Cref{sec:problem-setup}). Then we discuss the empirically effective structural prior of tabular data (\Cref{sec:data-structure}), and the proposed methodology for quantifying structural fidelity (\Cref{sec:structural-fidelity}). Next, we detail the conventional evaluation dimensions employed in TabStruct (\Cref{sec:conventional-eval-dim}). Finally, we introduce the benchmark datasets (\Cref{sec:benchmark-dataset}) and the benchmark generators (\Cref{sec:benchmark-generator}).

\subsection{Problem Setup}
\label{sec:problem-setup}
We address the task of tabular data generation. Let $\mathcal{D} \coloneqq \{(\vx^{(1)}, y^{(1)}), \dots, (\vx^{(N)}, y^{(N)})\}$ represent a labelled tabular dataset consisting of $N$ samples. For the $i$-th sample $\vx^{(i)}$, $x^{(i)}_{d}$ denotes its $d$-th feature, and $y^{(i)}$ denotes the corresponding target. To simplify notation, we refer to the training split of the full dataset $\mathcal{D}$ as the reference data, denoted by $\mathcal{D}_{\text{ref}}$. The synthetic data produced by tabular data generators is denoted by $\mathcal{D}_{\text{syn}}$. The evaluation of tabular generative models is conducted by assessing the quality of $\mathcal{D}_{\text{syn}}$ across multiple dimensions. We further illustrate the setup in~\Cref{appendix:data-splitting}.

\subsection{Tabular Data Structure}
\label{sec:data-structure}
The underlying structure of tabular data has long been an open research question~\citep{kitson2023survey, hollmann2025accurate, mullertransformers}. For other modalities like textual data, it is natural to characterise their structure as autoregressive, guided by human knowledge~\citep{yang2019xlnet}. Therefore, pretraining paradigms aligned with the autoregressive structure, such as next-token prediction~\citep{achiam2023gpt}, have proven successful in textual generative modelling. In contrast, heterogeneous tabular data does not naturally lend itself to human interpretation, making a structural prior for such data generally elusive.

\looseness-1
Recent studies~\citep{hollmann2025accurate, mullertransformers} on tabular foundation predictors have begun to shed light on the underlying structure of tabular data. \citet{hollmann2025accurate} introduces TabPFN, a tabular foundation predictor pretrained on 100 million ``synthetic'' tabular datasets. These datasets are ``synthetic'' because they do not incorporate real-world semantics: they are produced with randomly constructed structural causal models (SCM).
Remarkably, despite not being explicitly trained on any real-world dataset, TabPFN is able to outperform an ensemble of strong baseline predictors, which have been fine-tuned on each individual classification task. The exceptional performance of TabPFN suggests that the SCMs used to construct the pretraining datasets, despite lacking real-world semantics, effectively reflect the structural information encoded in real-world tabular data. 

However, it is important to note that this does not imply that SCMs can fully represent the underlying structures of all tabular data. Instead, TabPFN demonstrates that the causal relationships between features, as modelled by SCMs, act as an empirically effective structural prior for a great proportion of real-world tabular data. 

As the success of LLMs primarily stems from their ability to leverage the autoregressive nature of textual data, we argue that a robust tabular data generation process should be able to capture  the unique causal structures within the tabular data. More specifically, generating data aligned with the causal structures in reference data could provide valuable insights into the open research question of how to effectively leverage the structural information inherent in tabular data.

\subsection{Structural Fidelity}
\label{sec:structural-fidelity}
Using causal relationships as the structural prior for tabular data, we define the {\em structural fidelity} of a tabular generative model as the alignment between the causal structures in the reference data $\mathcal{D}_{\text{ref}}$ and the synthetic data $\mathcal{D}_{\text{syn}}$.
Following prior benchmarks on causal discovery and inference~\citep{spirtes2001causation, tu2024causality}, TabStruct evaluates structural fidelity at the level of the Markov equivalent class. At this level, causal structures are represented by completed partially directed acyclic graphs (CPDAGs). The causal structures of $\mathcal{D}_{\text{ref}}$ and $\mathcal{D}_{\text{syn}}$ are considered equivalent as long as they encode the same set of conditional independence relationships between features.

\begin{figure}[!t]
    \centering
    \includegraphics[width=\textwidth]{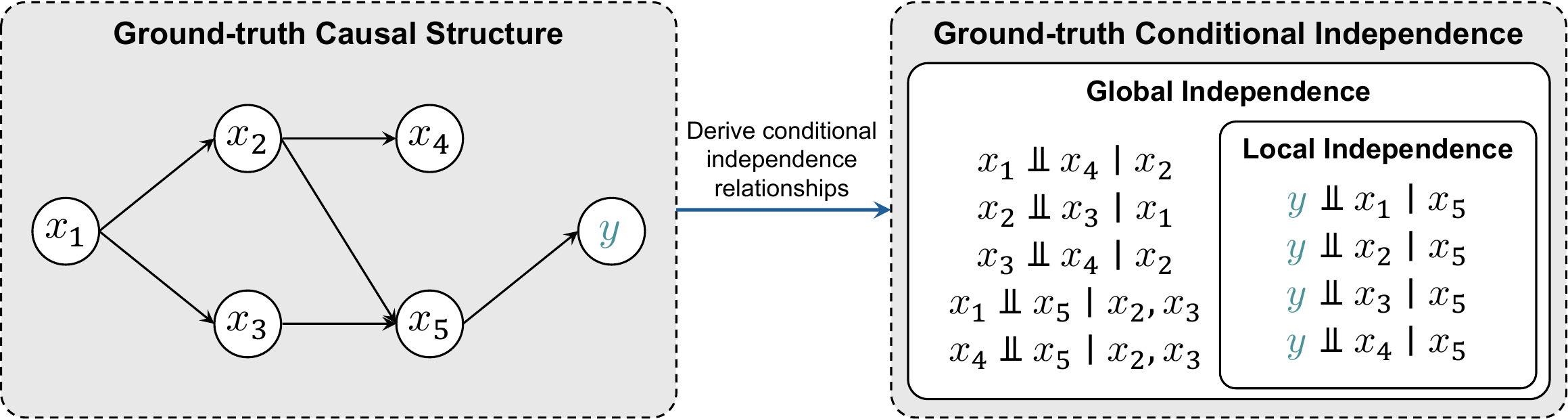}
    \caption{\textbf{An illustrative example for the quantification of structural fidelity.} Given the ground-truth causal structure, we first derive the conditional independence relationships between features. These relationships are then divided into two levels of granularity: global and local. The global set encompasses all conditional independence relationships across the entire feature set, whereas the local set includes only those relationships that are directly relevant to the {\color[HTML]{008080} {target variable $y$}}. Next, we apply conditional independence tests on $\mathcal{D}_{\text{syn}}$ to examine the alignment of conditional independence relationships between features.}
\label{fig:ci-test}
\end{figure}

\paragraph{Fine-grained quantification of structural fidelity.}
Given the ground-truth causal structure of $\mathcal{D}_{\text{ref}}$, we can derive all conditional independence relationships between features programmatically (\Cref{fig:ci-test}). These conditional independence relationships are then tested on $\mathcal{D}_{\text{syn}}$ to investigate whether the synthetic data exhibits a Markov equivalent causal structure to the reference data. For each pair of features, the conditional independence test is formulated as a binary classification task, where 1 indicates independence and 0 indicates dependence. The balanced accuracy of the conditional independence tests on $\mathcal{D}_{\text{syn}}$ is then computed in order to quantify structural fidelity.

To provide a more fine-grained assessment of structural fidelity, we decompose structural fidelity into two complementary metrics: \textit{global independence} and \textit{local independence}. As illustrated in~\Cref{fig:ci-test}, global independence evaluates all conditional independence relationships in the dataset, whereas local independence focuses only on those relationships relevant to the target variable $y$. Intuitively, local independence assesses how well the generator models the relationships between the target and the features, while global independence provides a comprehensive evaluation of the generator's ability to capture the overall structure of tabular data.

\paragraph{Rationales for CPDAG-level evaluation.}
TabStruct does not evaluate causal fidelity at the directed acyclic graph (DAG) level, as this would require an additional causal discovery method to determine the causal directions between features. 
Recovering causal directions is an inherently challenging task, and no existing causal discovery methods can guarantee perfect identification of causal directions~\citep{zanga2022survey, kaddour2022causal}. 
This limitation is further illustrated in \Cref{sec:exp}. Additionally, evaluating at the DAG level can introduce biases, as the results depend on the specific causal discovery method used. This issue is similar to the key limitation of ``downstream utility'', which is inherently biased by the choice of downstream tasks and predictors. To address this, TabStruct evaluates causal structures at the CPDAG level, reducing the risks associated with inaccurate or biased identification of causal directions.

\subsection{Conventional Evaluation Approaches}
\label{sec:conventional-eval-dim}
\looseness-1
\textbf{Density estimation} evaluates the mismatch between the marginal (i.e., low-order) or joint (i.e.,~high-order) distributions of reference and synthetic data~\citep{hansen2023reimagining}. A generator can trivially achieve high performance on low-order metrics by independently sampling from each feature's marginal distribution. While high-order metrics measure sample-level similarity, they still fail to explicitly demonstrate whether the synthetic data presents the same causal structures as reference data.

Following prior studies~\citep{hansen2023reimagining, shi2024tabdiff, zhangmixed}, we evaluate density estimation using four metrics of two categories:  
(i)~Low-order: \textit{Shape} and \textit{Trend}~\citep{wust2011sdmetrics}. Shape measures the synthetic data's ability to replicate each column's marginal density. Trend assesses its capacity to capture correlations between different columns.
(ii)~High-order: \mbox{\textit{$\alpha$-precision}} and \mbox{\textit{$\beta$-recall}}~\citep{alaa2022faithful}. \textit{$\alpha$-precision} quantifies the similarity between the reference and synthetic data, and \textit{$\beta$-recall} assesses the diversity of the synthetic data.

\textbf{Downstream utility} measures the performance gap when substituting reference data with synthetic data in downstream tasks. This metric is inherently task-specific and susceptible to bias from the choice of downstream models and tasks. A parallel can be drawn to image generation, where Mixup~\citep{psaroudakis2022mixaugment} augments training data with synthetic samples by interpolating between real samples. While Mixup improves downstream performance, it disrupts the spatial structure of images, resulting in synthetic samples that are generally visually unrealistic~\citep{mumuni2022data}. This example shows that downstream utility, while useful for specific tasks, cannot serve as a holistic measure of a tabular data generator.

\looseness-1
For all downstream tasks, we adopt the ``train-on-synthetic, test-on-real'' strategy~\citep{xu2019modeling}. To mitigate the bias from downstream models, we evaluate downstream utility by averaging the performance of six representative downstream predictors, including three standard baselines: Logistic Regression (LR)~\citep{cox1958regression}, KNN~\citep{fix1985discriminatory} and MLP~\citep{gorishniy2021revisiting}; two tree-based methods: Random Forest (RF)~\citep{breiman2001random} and XGBoost~\citep{chen2016xgboost}; and a PFN method: TabPFN~\citep{hollmann2025accurate}.

\textbf{Privacy preservation} primarily focuses on the trade-off between specific downstream tasks and privacy leakage~\citep{margeloiutabebm}. Similar to downstream utility, this dimension is also highly task-dependent, making it susceptible to bias, and limiting its ability to provide a comprehensive assessment of the capability of tabular generative models. 

We measure privacy preservation using two metrics:  
(i)~\textit{median Distance to Closest Record} (DCR)~\citep{zhao2021ctab}, where a higher DCR indicates that synthetic data is less likely to be directly copied from the reference data;  
(ii)~\textit{Authenticity}~\citep{alaa2022faithful}, where a higher score indicates that the generated samples are less likely to be mere replicas of the reference data.  

\subsection{Benchmark Datasets with Ground Truth Causal Structures}
\label{sec:benchmark-dataset}
\looseness-1
To accurately quantify structural fidelity, the reference data should be paired with ground-truth causal structures. Therefore, we construct benchmark datasets by leveraging structural causal models (SCMs) that have been validated by human experts~\citep{scutari2011bnlearn}. Human validation ensures that the causal structures are realistic, increasing the likelihood that TabStruct's benchmark results can generalise to other real-world datasets without known causal structures (i.e., where structural fidelity cannot be directly evaluated). We note that this is a core distinction between TabStruct and prior studies~\citep{tu2024causality, hollmann2025accurate}: instead of relying on datasets without real-world semantics, TabStruct utilises reference data with expert-validated, realistic causal structures and mixed feature types.

We outline the process of building the reference datasets as follows. Firstly, we use ground-truth SCMs with realistic and expert-validated structures. Secondly, we perform prior sampling on these SCMs: root nodes are randomly initialised, and their values are propagated through the causal graph. A single sample is generated by recording the node values after propagation, with each propagation producing one sample. Thirdly, this process is repeated until sufficient samples are obtained. By following this procedure, we construct full datasets $\mathcal{D}$ with accessible and well-defined causal structures. We include both classification (\Cref{tab:dataset-classification-structure}) and regression (\Cref{tab:dataset-regression-structure}) datasets, and the detailed descriptions are in \Cref{appendix:dataset}.

\subsection{Benchmark Generators}
\label{sec:benchmark-generator}
\looseness-1
TabStruct includes nine existing tabular data generation methods of eight different categories: 
(i)~a standard interpolation method SMOTE~\citep{chawla2002smote}; 
(ii)~a structure learning method Bayesian Network~\citep{qian2024synthcity}; 
(iii)~two Variational Autoencoders (VAE) based methods TVAE~\citep{xu2019modeling} and GOGGLE~\citep{liu2023goggle}; 
(iv)~a Generative Adversarial Networks (GAN) method CTGAN~\citep{xu2019modeling}; 
(v)~a normalising flow model Neural Spine Flows (NFLOW)~\citep{durkan2019neural}; 
(vi)~a diffusion model TabDDPM~\citep{kotelnikov2023tabddpm}; 
(vii)~a tree-based method Adversarial Random Forests (ARF)~\citep{watson2023adversarial}; and 
(viii)~a Large Language Model (LLM) based method GReaT~\citep{borisovlanguage}. 
In addition, we include $\mathcal{D}_{\text{ref}}$, where the reference data is directly used for evaluation. We provide full implementation details of benchmark generators in~\Cref{appendix:implementation-generator}.

\section{Experiments}
\label{sec:exp}

\paragraph{Experimental setup.}
For each dataset of $N$ samples, we first split it into train and test sets (80\% train and 20\% test). We further split the train set into a training split ($\mathcal{D}_{\text{ref}}$) and a validation split (90\% training and 10\% validation). For classification datasets, stratification is preserved during data splitting. We provide detailed descriptions of data splitting in \Cref{appendix:reproducibility}. We repeat the splitting 10 times, summing up to 10 runs per dataset.
All benchmark generators are trained on $\mathcal{D}_{\text{ref}}$, and each generator produces a synthetic dataset with $N_{\text{syn}}$ samples. For classification, the synthetic data preserves the stratification of reference data. Since a small $N_{\text{syn}}$ may not yield robust results of model performance~\citep{margeloiutabebm}, we conduct a proof-of-concept experiment (see \Cref{appendix:syn-sample-size} for more details) and empirically set $N_{\text{syn}} = 3N_{\text{ref}}$ as the saturation point where further increases in $N_{\text{syn}}$ have negligible impact on evaluation results.

\paragraph{Aggregation of evaluation results.}
The reported results are averaged by default over 10 runs on the test sets. When aggregating results across datasets, we use the average distance to the minimum (ADTM) metric via affine renormalisation between the top-performing and worse-performing models~\citep{grinsztajn2022tree, mcelfresh2024neural, hollmann2025accurate, margeloiutabebm, jiangprotogate}. To aggregate different metrics within the same evaluation dimension, we compute their average.
For downstream utility, evaluation results are averaged over six downstream predictors to mitigate the bias from specific predictors.

\subsection{Generator Performance in Learning Tabular Data Structure}

\begin{table}[!t]
\setlength{\tabcolsep}{3pt}
\centering
\caption{\textbf{Benchmark results of nine tabular data generators on seven datasets with varying feature scales.} The results are grouped based on the tasks. For each group, we report the normalised mean $\pm$ std metric values across datasets. We also highlight the {\color[HTML]{008080} \textbf{First}}, {\color[HTML]{7030A0} \textbf{Second}} and {\color[HTML]{C65911} \textbf{Third}} best performances for each metric. Existing tabular generative models, including advanced neural networks, struggle to accurately capture the underlying structure of tabular data.
\label{tab:summary-top3}}
\resizebox{\textwidth}{!}{
\begin{tabular}{l|rrrr|rr|rr|rr}

\toprule

\multirow{2}{*}{\textbf{Generator}} & \multicolumn{4}{c|}{\textbf{Density   Estimation}}                             & \multicolumn{2}{c|}{\textbf{Downstream   Utility}}                             & \multicolumn{2}{c|}{\textbf{Privacy Preservation}} & \multicolumn{2}{c}{\textbf{Structural Fidelity}} \\

& Shape $\uparrow$ & Trend $\uparrow$ & $\alpha$-precision $\uparrow$ & $\beta$-recall $\uparrow$ & Accuracy $\uparrow$ & RMSE $\downarrow$ & DCR $\uparrow$ & Authenticity $\uparrow$ & Local independence $\uparrow$ & Global independence $\uparrow$ \\

\midrule
\rowcolor{Gainsboro!60}
\multicolumn{11}{c}{\textbf{Classification tasks}} \\
\midrule

$\mathcal{D}_{\text{ref}}$ &1.00$_{\pm\text{0.00}}$ & 1.00$_{\pm\text{0.00}}$ & 1.00$_{\pm\text{0.00}}$ & 1.00$_{\pm\text{0.00}}$ & 100.00$_{\pm\text{0.00}}$ & $-$ & 0.00$_{\pm\text{0.00}}$ & 0.00$_{\pm\text{0.00}}$ & 100.00$_{\pm\text{0.00}}$ & 100.00$_{\pm\text{0.00}}$ \\

\midrule

SMOTE & {\color[HTML]{C65911} \textbf{0.94}}$_{\pm\text{0.00}}$ & {\color[HTML]{008080} \textbf{0.93}}$_{\pm\text{0.00}}$ & 0.82$_{\pm\text{0.01}}$ & {\color[HTML]{008080} \textbf{1.00}}$_{\pm\text{0.00}}$ & {\color[HTML]{008080} \textbf{95.71}}$_{\pm\text{1.54}}$ & $-$ & 0.11$_{\pm\text{0.01}}$ & 0.05$_{\pm\text{0.01}}$ & {\color[HTML]{008080} \textbf{74.02}}$_{\pm\text{3.13}}$ & 35.39$_{\pm\text{0.85}}$ \\

BN & {\color[HTML]{008080} \textbf{0.97}}$_{\pm\text{0.00}}$ & {\color[HTML]{7030A0} \textbf{0.91}}$_{\pm\text{0.00}}$ & {\color[HTML]{008080} \textbf{0.95}}$_{\pm\text{0.01}}$ & {\color[HTML]{7030A0} \textbf{0.75}}$_{\pm\text{0.01}}$ & {\color[HTML]{C65911} \textbf{89.88}}$_{\pm\text{1.01}}$ & $-$ & 0.79$_{\pm\text{0.03}}$ & 0.50$_{\pm\text{0.01}}$ & 35.49$_{\pm\text{3.42}}$ & 45.31$_{\pm\text{0.79}}$ \\

TVAE & 0.83$_{\pm\text{0.00}}$ & 0.75$_{\pm\text{0.00}}$ & 0.70$_{\pm\text{0.02}}$ & {\color[HTML]{C65911} \textbf{0.66}}$_{\pm\text{0.02}}$ & {\color[HTML]{7030A0} \textbf{94.57}}$_{\pm\text{1.50}}$ & $-$ & {\color[HTML]{7030A0} \textbf{0.94}}$_{\pm\text{0.04}}$ & 0.60$_{\pm\text{0.02}}$ & {\color[HTML]{7030A0} \textbf{65.96}}$_{\pm\text{3.64}}$ & {\color[HTML]{008080} \textbf{64.29}}$_{\pm\text{0.77}}$ \\

GOGGLE & 0.08$_{\pm\text{0.02}}$ & 0.05$_{\pm\text{0.01}}$ & 0.01$_{\pm\text{0.01}}$ & 0.24$_{\pm\text{0.02}}$ & 18.57$_{\pm\text{1.16}}$ & $-$ & {\color[HTML]{008080} \textbf{0.98}}$_{\pm\text{0.03}}$ & {\color[HTML]{7030A0} \textbf{0.86}}$_{\pm\text{0.01}}$ & 0.00$_{\pm\text{0.00}}$ & 0.00$_{\pm\text{0.00}}$ \\

CTGAN & 0.72$_{\pm\text{0.02}}$ & 0.74$_{\pm\text{0.02}}$ & {\color[HTML]{C65911} \textbf{0.89}}$_{\pm\text{0.04}}$ & 0.52$_{\pm\text{0.06}}$ & 76.74$_{\pm\text{2.80}}$ & $-$ & 0.82$_{\pm\text{0.03}}$ & 0.74$_{\pm\text{0.05}}$ & {\color[HTML]{C65911} \textbf{55.58}}$_{\pm\text{0.47}}$ & {\color[HTML]{7030A0} \textbf{50.95}}$_{\pm\text{1.02}}$ \\

NFlow & 0.73$_{\pm\text{0.01}}$ & 0.66$_{\pm\text{0.01}}$ & 0.79$_{\pm\text{0.03}}$ & 0.20$_{\pm\text{0.03}}$ & 23.79$_{\pm\text{3.02}}$ & $-$ & 0.84$_{\pm\text{0.04}}$ & {\color[HTML]{008080} \textbf{0.92}}$_{\pm\text{0.01}}$ & 20.88$_{\pm\text{4.40}}$ & 40.74$_{\pm\text{1.14}}$ \\

TabDDPM & 0.39$_{\pm\text{0.01}}$ & 0.37$_{\pm\text{0.01}}$ & 0.24$_{\pm\text{0.01}}$ & 0.22$_{\pm\text{0.01}}$ & 33.90$_{\pm\text{1.07}}$ & $-$ & 0.77$_{\pm\text{0.04}}$ & {\color[HTML]{C65911} \textbf{0.83}}$_{\pm\text{0.01}}$ & 5.63$_{\pm\text{2.62}}$ & 23.69$_{\pm\text{0.68}}$ \\

ARF & {\color[HTML]{7030A0} \textbf{0.97}}$_{\pm\text{0.00}}$ & {\color[HTML]{C65911} \textbf{0.90}}$_{\pm\text{0.00}}$ & {\color[HTML]{7030A0} \textbf{0.92}}$_{\pm\text{0.01}}$ & 0.61$_{\pm\text{0.02}}$ & 59.21$_{\pm\text{2.15}}$ & $-$ & {\color[HTML]{C65911} \textbf{0.85}}$_{\pm\text{0.03}}$ & 0.65$_{\pm\text{0.01}}$ & 32.59$_{\pm\text{3.72}}$ & {\color[HTML]{C65911} \textbf{46.40}}$_{\pm\text{0.94}}$ \\

GReaT & 0.74$_{\pm\text{0.01}}$ & 0.71$_{\pm\text{0.01}}$ & 0.65$_{\pm\text{0.02}}$ & 0.56$_{\pm\text{0.02}}$ & 48.34$_{\pm\text{1.57}}$ & $-$ & 0.69$_{\pm\text{0.03}}$ & 0.62$_{\pm\text{0.01}}$ & 38.56$_{\pm\text{3.15}}$ & 45.20$_{\pm\text{0.77}}$ \\



\midrule
\rowcolor{Gainsboro!60}
\multicolumn{11}{c}{\textbf{Regression datasets}} \\
\midrule

$\mathcal{D}_{\text{ref}}$ &1.00$_{\pm\text{0.00}}$ & 1.00$_{\pm\text{0.00}}$ & 1.00$_{\pm\text{0.00}}$ & 1.00$_{\pm\text{0.00}}$ & $-$ & 0.00$_{\pm\text{0.01}}$ & 0.00$_{\pm\text{0.00}}$ & 0.00$_{\pm\text{0.00}}$ & 100.00$_{\pm\text{0.00}}$ & 100.00$_{\pm\text{0.00}}$ \\

\midrule

SMOTE & {\color[HTML]{7030A0} \textbf{0.85}}$_{\pm\text{0.00}}$ & {\color[HTML]{008080} \textbf{0.89}}$_{\pm\text{0.00}}$ & 0.71$_{\pm\text{0.00}}$ & {\color[HTML]{008080} \textbf{0.96}}$_{\pm\text{0.00}}$ & $-$ & {\color[HTML]{C65911} \textbf{0.16}}$_{\pm\text{0.02}}$ & 0.29$_{\pm\text{0.01}}$ & 0.08$_{\pm\text{0.01}}$ & {\color[HTML]{7030A0} \textbf{49.66}}$_{\pm\text{6.67}}$ & {\color[HTML]{7030A0} \textbf{69.21}}$_{\pm\text{5.36}}$ \\

BN & {\color[HTML]{008080} \textbf{0.86}}$_{\pm\text{0.00}}$ & {\color[HTML]{C65911} \textbf{0.77}}$_{\pm\text{0.00}}$ & {\color[HTML]{7030A0} \textbf{0.90}}$_{\pm\text{0.01}}$ & {\color[HTML]{7030A0} \textbf{0.73}}$_{\pm\text{0.01}}$ & $-$ & {\color[HTML]{7030A0} \textbf{0.10}}$_{\pm\text{0.02}}$ & 0.22$_{\pm\text{0.01}}$ & 0.30$_{\pm\text{0.01}}$ & {\color[HTML]{008080} \textbf{71.05}}$_{\pm\text{5.00}}$ & {\color[HTML]{008080} \textbf{77.51}}$_{\pm\text{2.96}}$ \\

TVAE & 0.70$_{\pm\text{0.01}}$ & 0.61$_{\pm\text{0.01}}$ & 0.61$_{\pm\text{0.03}}$ & 0.63$_{\pm\text{0.02}}$ & $-$ & 0.20$_{\pm\text{0.27}}$ & {\color[HTML]{7030A0} \textbf{0.66}}$_{\pm\text{0.04}}$ & 0.50$_{\pm\text{0.02}}$ & {\color[HTML]{C65911} \textbf{44.10}}$_{\pm\text{4.15}}$ & {\color[HTML]{C65911} \textbf{52.52}}$_{\pm\text{2.31}}$ \\

GOGGLE & 0.30$_{\pm\text{0.06}}$ & 0.24$_{\pm\text{0.01}}$ & 0.30$_{\pm\text{0.09}}$ & 0.28$_{\pm\text{0.03}}$ & $-$ & 0.83$_{\pm\text{0.29}}$ & 0.32$_{\pm\text{0.06}}$ & {\color[HTML]{7030A0} \textbf{0.80}}$_{\pm\text{0.02}}$ & 21.31$_{\pm\text{1.47}}$ & 22.33$_{\pm\text{0.74}}$ \\

CTGAN & 0.54$_{\pm\text{0.04}}$ & 0.51$_{\pm\text{0.02}}$ & {\color[HTML]{C65911} \textbf{0.73}}$_{\pm\text{0.09}}$ & 0.44$_{\pm\text{0.08}}$ & $-$ & 0.37$_{\pm\text{0.55}}$ & 0.31$_{\pm\text{0.03}}$ & {\color[HTML]{C65911} \textbf{0.76}}$_{\pm\text{0.05}}$ & 9.21$_{\pm\text{6.09}}$ & 11.96$_{\pm\text{4.11}}$ \\

NFlow & 0.74$_{\pm\text{0.01}}$ & 0.62$_{\pm\text{0.01}}$ & 0.67$_{\pm\text{0.04}}$ & 0.52$_{\pm\text{0.05}}$ & $-$ & 0.31$_{\pm\text{0.08}}$ & {\color[HTML]{C65911} \textbf{0.58}}$_{\pm\text{0.04}}$ & 0.72$_{\pm\text{0.04}}$ & 37.62$_{\pm\text{3.62}}$ & 23.68$_{\pm\text{3.70}}$ \\

TabDDPM & 0.20$_{\pm\text{0.02}}$ & 0.25$_{\pm\text{0.01}}$ & 0.31$_{\pm\text{0.00}}$ & 0.18$_{\pm\text{0.01}}$ & $-$ & {\color[HTML]{008080} \textbf{0.02}}$_{\pm\text{0.03}}$ & {\color[HTML]{008080} \textbf{0.68}}$_{\pm\text{0.03}}$ & {\color[HTML]{008080} \textbf{0.87}}$_{\pm\text{0.00}}$ & 38.97$_{\pm\text{0.72}}$ & 10.20$_{\pm\text{5.33}}$ \\

ARF & {\color[HTML]{C65911} \textbf{0.84}}$_{\pm\text{0.00}}$ & {\color[HTML]{7030A0} \textbf{0.79}}$_{\pm\text{0.00}}$ & {\color[HTML]{008080} \textbf{0.94}}$_{\pm\text{0.01}}$ & 0.53$_{\pm\text{0.02}}$ & $-$ & 0.18$_{\pm\text{0.26}}$ & 0.36$_{\pm\text{0.01}}$ & 0.66$_{\pm\text{0.02}}$ & 33.64$_{\pm\text{3.50}}$ & 31.56$_{\pm\text{2.32}}$ \\

GReaT & 0.67$_{\pm\text{0.01}}$ & 0.67$_{\pm\text{0.01}}$ & 0.69$_{\pm\text{0.03}}$ & {\color[HTML]{C65911} \textbf{0.64}}$_{\pm\text{0.03}}$ & $-$ & 0.21$_{\pm\text{0.25}}$ & 0.39$_{\pm\text{0.03}}$ & 0.51$_{\pm\text{0.02}}$ & 38.42$_{\pm\text{5.09}}$ & 38.66$_{\pm\text{3.38}}$ \\  
\bottomrule
\end{tabular}
}
\end{table}
%
%
\begin{table}[t!]
\centering
\caption{\textbf{Correlation between ranks of different metrics.} The relatively higher correlation between Accuracy/RMSE and local independence demonstrates that a generator can achieve high downstream utility by prioritising the conditional independence relationships directly relevant to the target variable while overlooking the global structure.
\label{tab:summary-rank-correlation}}
\begin{tabular}{lrr}
\toprule
                                          & Local independence & Global independence \\
\midrule
Accuracy & \textbf{0.90}             & 0.57              \\
RMSE & \textbf{0.77}             & 0.33     \\
\bottomrule
\end{tabular}%
\end{table}
\paragraph{Downstream utility is not the golden standard for tabular generative modelling.}
In prior studies (\Cref{appendix:literature-review}), downstream utility is often considered as the core evaluation dimension. From this perspective, a generator is considered effective if its synthetic data achieves high performance in downstream tasks. However, as discussed in \Cref{sec:conventional-eval-dim}, downstream utility inherently biases evaluation towards relationships between the target variable and the features, thus overlooking the relationships between features.  
\Cref{tab:summary-top3} and \Cref{tab:summary-rank-correlation} quantitatively demonstrate this limitation. The rankings of downstream utility are strongly correlated with local independence but exhibit much weaker correlation with global independence. This indicates that a generator can achieve high downstream utility by prioritising local independence at the expense of global independence. For instance, SMOTE achieves the highest downstream utility and local independence in classification tasks, but it performs poorly in global independence. This suggests that SMOTE focuses narrowly on structures relevant to the target variable while neglecting inter-feature relationships. 
Therefore, a generator should not be deemed effective solely based on downstream utility, as it can overlook the broader structural information encoded in the data.

\paragraph{Structural fidelity presents consistent challenges across tasks and generators.}  
\label{sec:exp-structural-fidelity}
\Cref{tab:summary-top3} shows a notable gap in structural fidelity between reference data ($\mathcal{D}_{\text{ref}}$) and synthetic data ($\mathcal{D}_{\text{syn}}$) across both classification and regression tasks. For instance, in classification tasks, the highest local independence achieved is 74.02\% (SMOTE), indicating the smallest performance gap relative to $\mathcal{D}_{\text{ref}}$ is over~25\%. Global independence shows an even large performance gap of 35\% between $\mathcal{D}_{\text{ref}}$ and $\mathcal{D}_{\text{syn}}$. In contrast, the smallest gaps between $\mathcal{D}_{\text{ref}}$ and $\mathcal{D}_{\text{syn}}$ in statistical fidelity and downstream utility remain consistently below 10\%. The underperformance in structural fidelity also exists in regression datasets.
These results underline the consistent challenges faced by existing tabular generative models in capturing the underlying structure of tabular data.  

\begin{figure}[t!]
    \centering
    \subfloat{\includegraphics[width=0.33\textwidth]{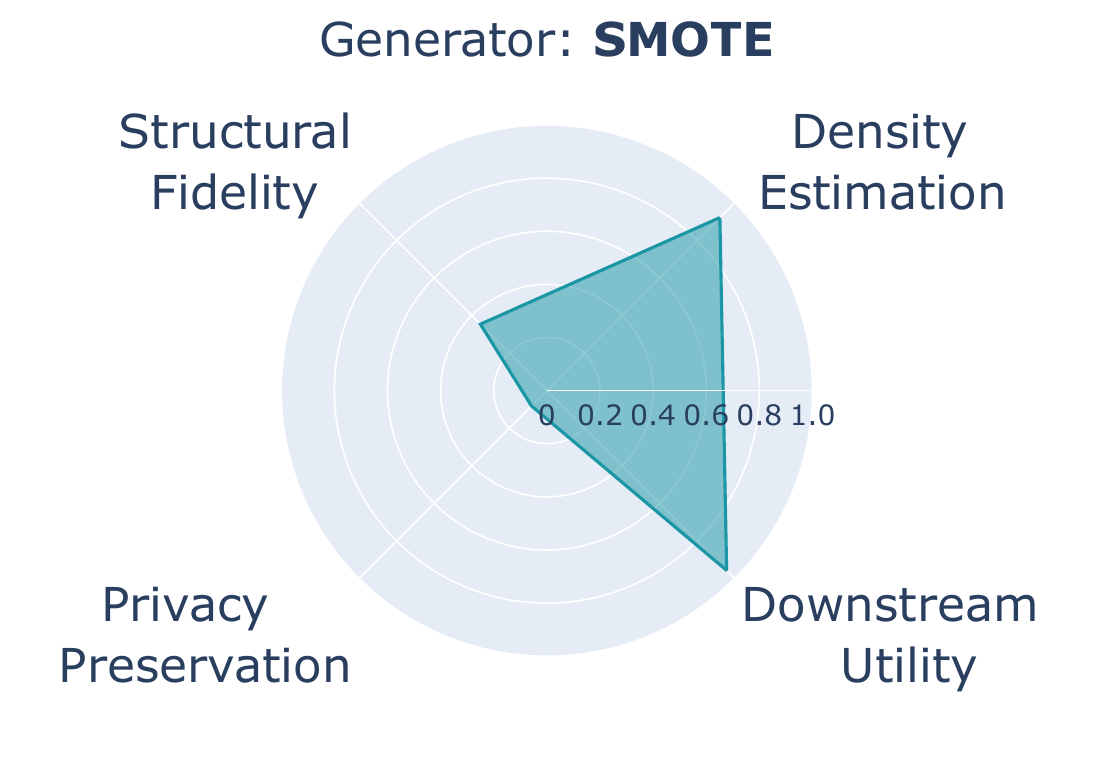}}
    \hfill
    \subfloat{\includegraphics[width=0.33\textwidth]{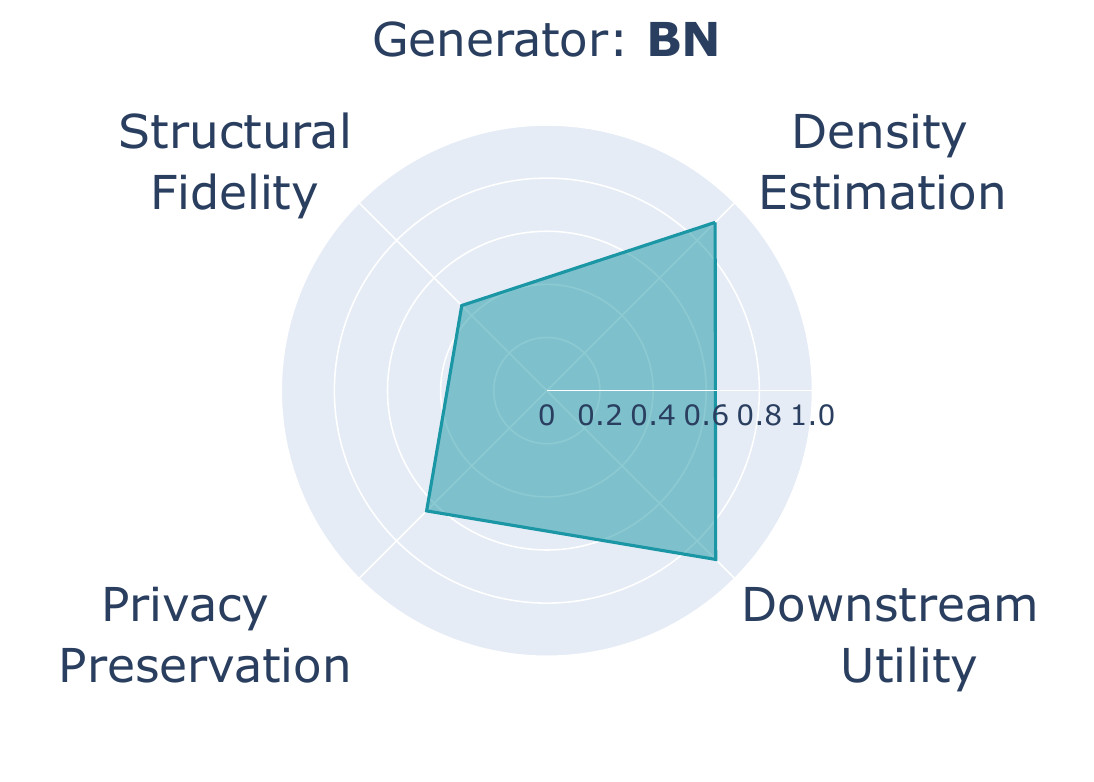}}
    \hfill
    \subfloat{\includegraphics[width=0.33\textwidth]{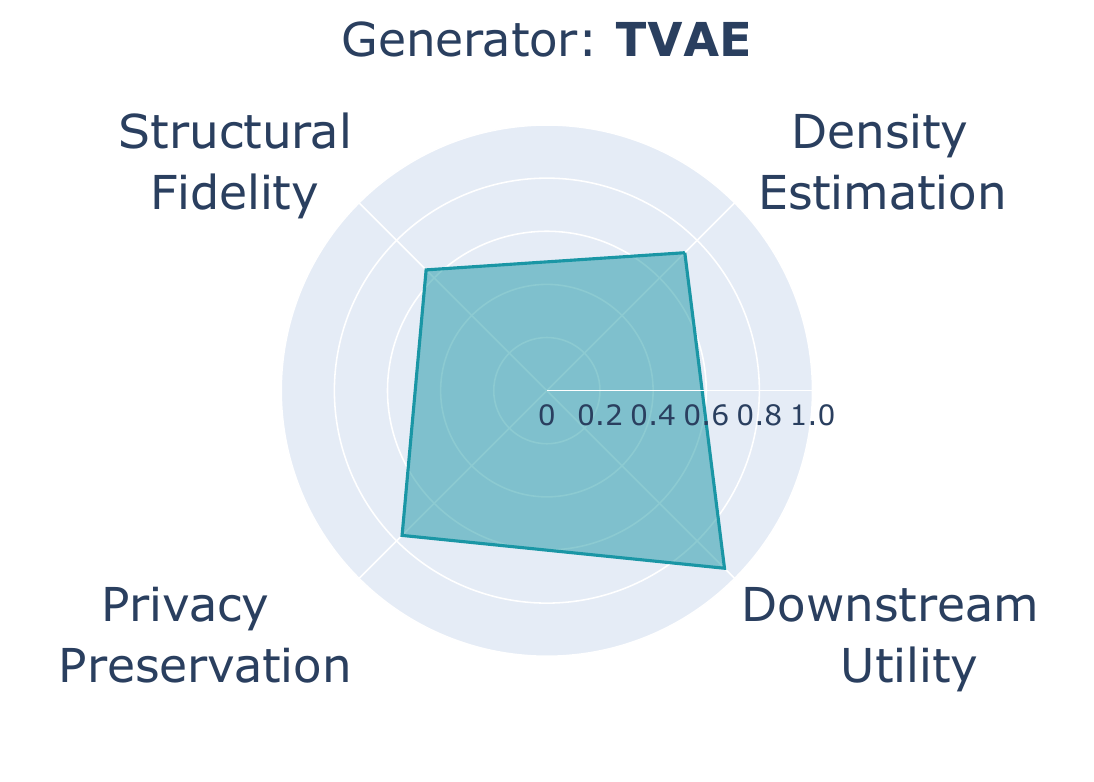}}
    \\
    \subfloat{\includegraphics[width=0.33\textwidth]{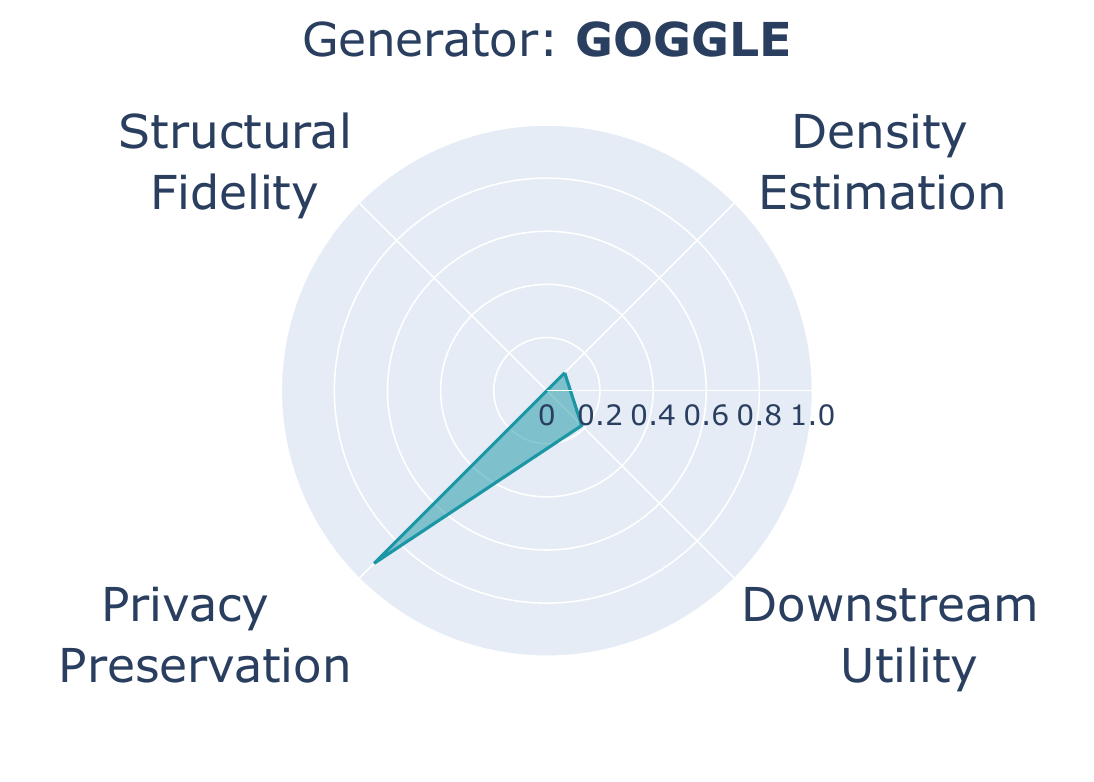}}
    \hfill
    \subfloat{\includegraphics[width=0.33\textwidth]{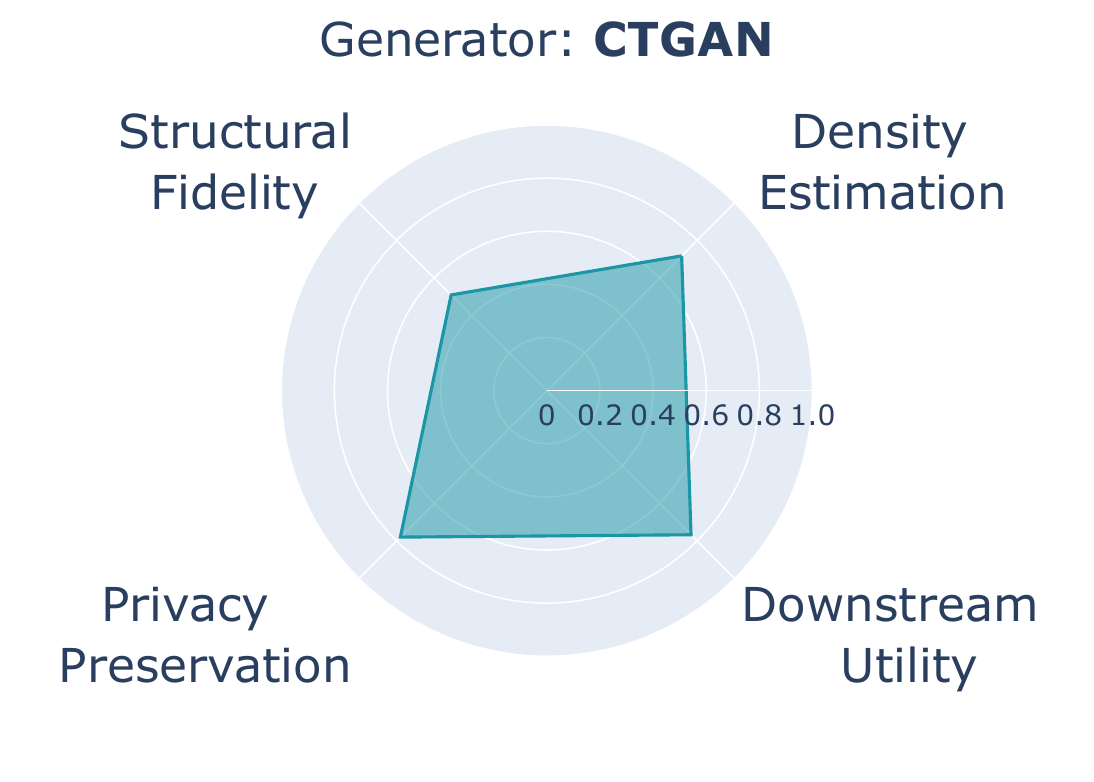}}
    \hfill
    \subfloat{\includegraphics[width=0.33\textwidth]{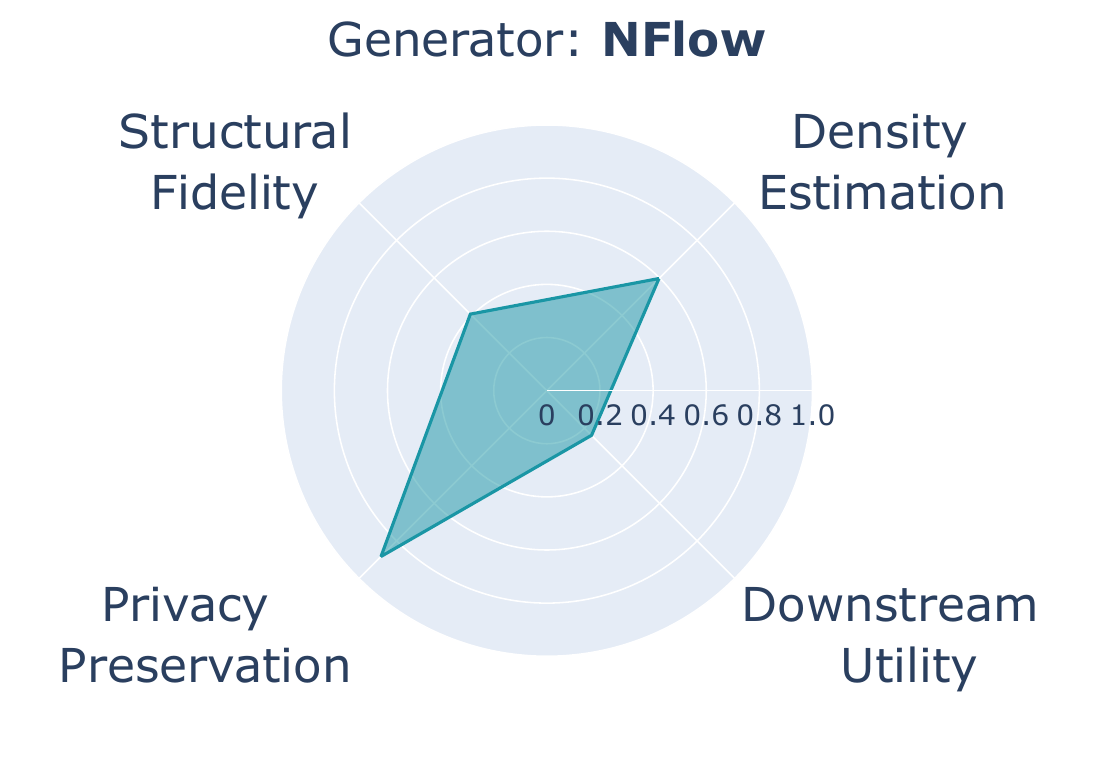}}
    \\
    \subfloat{\includegraphics[width=0.33\textwidth]{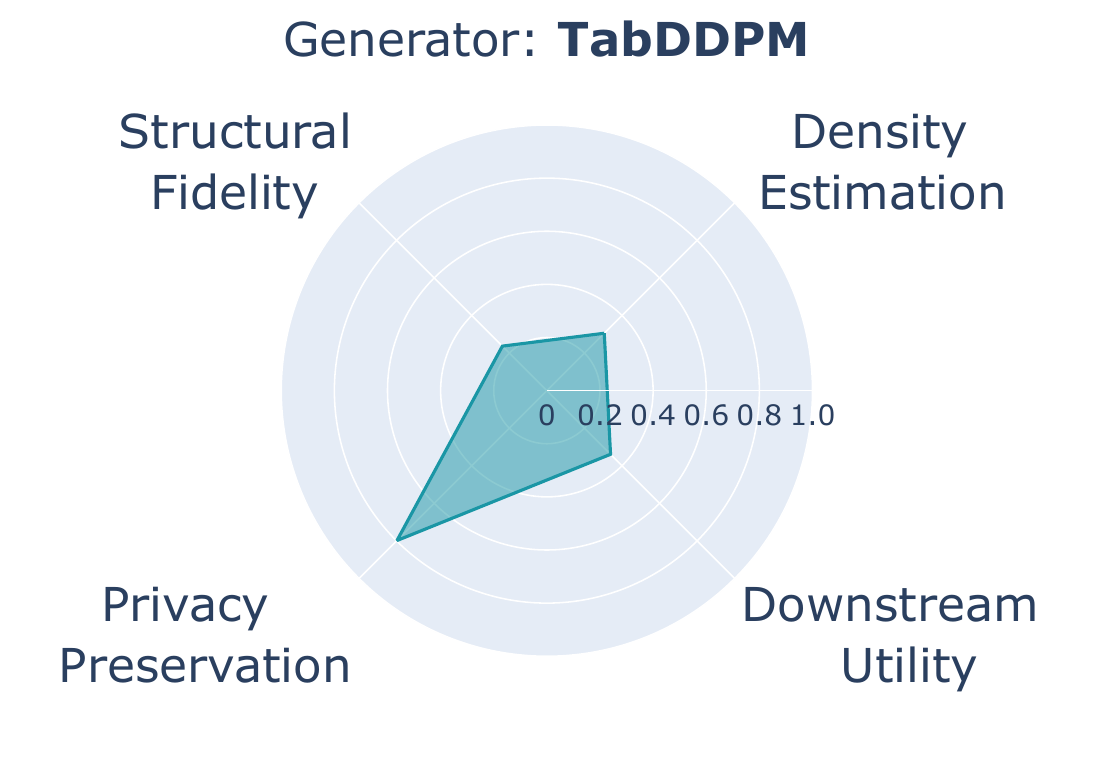}}
    \hfill
    \subfloat{\includegraphics[width=0.33\textwidth]{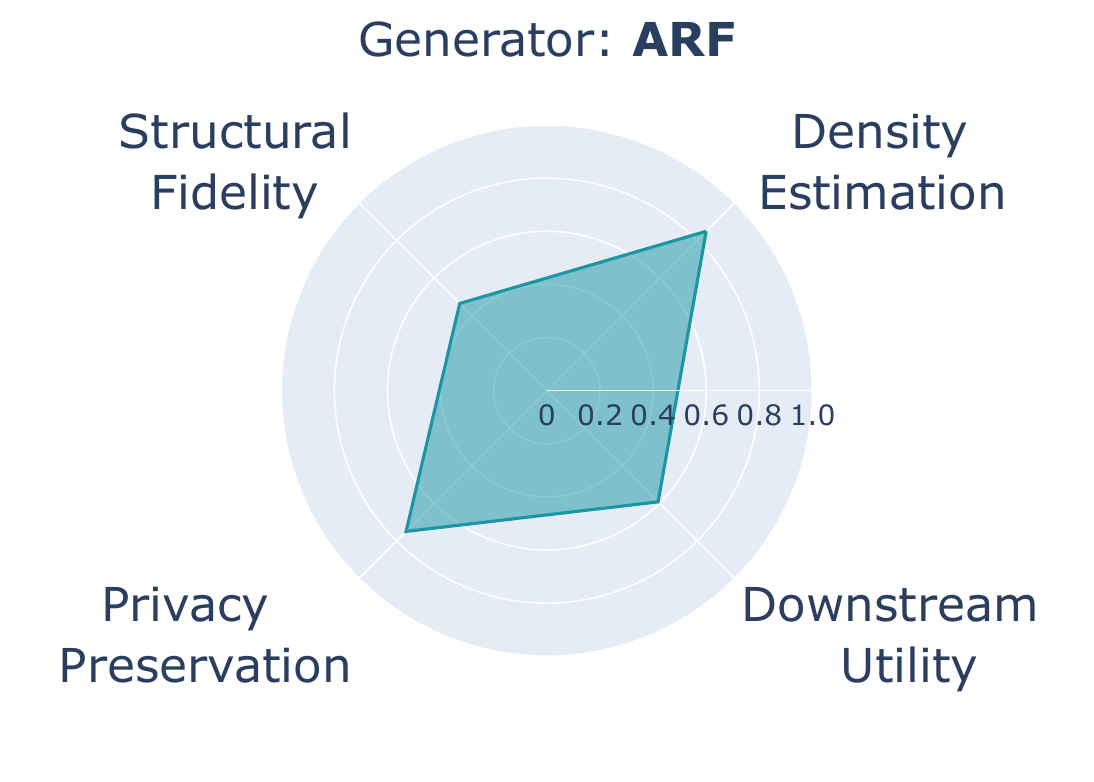}}
    \hfill
    \subfloat{\includegraphics[width=0.33\textwidth]{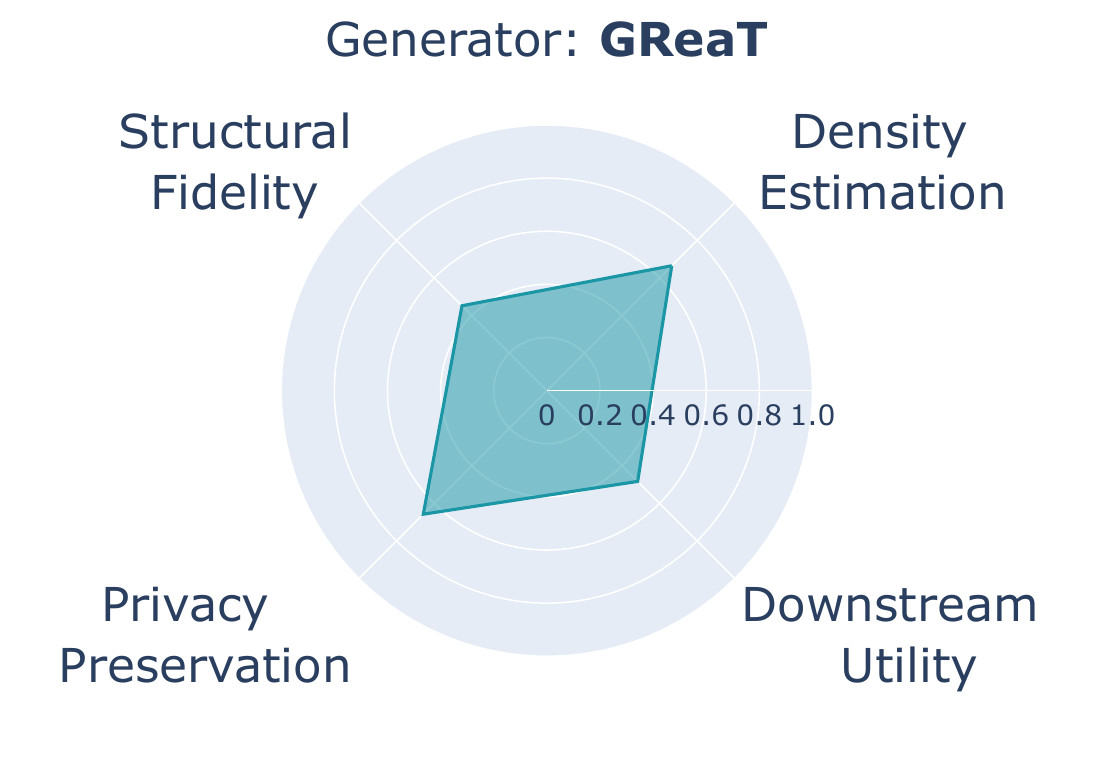}}
    \\  
    \caption{\textbf{Summarised comparison of nine tabular data generators across four evaluation dimensions.} The results reveal that excelling in conventional evaluation dimensions does not ensure the model's ability to capture the underlying data structure. Learning the underlying data structure remains challenging for tabular generative modelling.\label{fig:radar-chart}}
\end{figure}

\paragraph{Existing structure learning methods struggle with tabular data generation.}  
While Bayesian Network (BN) exhibit relatively strong performance in structural fidelity, their success is unsurprising -- the reference datasets are constructed with SCMs that perfectly align with the required assumptions of the causal discovery methods employed in BN (i.e., causal Markov assumption, causal sufficiency and causal faithfulness). 
Despite this advantage, the gap between $\mathcal{D}_{\text{ref}}$ and $\mathcal{D}_{\text{syn}}$ remains notable for BN. For instance, in classification tasks, the global independence gap exceeds 50\% compared to $\mathcal{D}_{\text{ref}}$. This demonstrates the limitations of existing structure learning methods in recovering perfect causal structures from observed data alone. Such findings are consistent with previous research~\citep{tu2024causality}, which reveals that current causal discovery methods struggle with datasets containing more than 10 features. 
In TabStruct, we employ realistic SCMs, with the number of features ranging from 7 to 223. Consequently, BNs perform less effectively despite having an objective function for explicit structure learning. This further justifies our choice to evaluate structural fidelity at the CPDAG level rather than the DAG level.

\vspace{-1mm}
\paragraph{Baseline models can outperform complex models in structural fidelity.}  
Interestingly, simple baseline models such as SMOTE and TVAE exhibit competitive performance in structural fidelity. For global independence, \Cref{tab:summary-top3} shows that TVAE consistently ranks among the top-3 across both classification and regression datasets. We note that TVAE does not possess explicit advantages as a structure learning method, indicating that variational autoencoders remain effective models for capturing feature relationships in tabular data.

\vspace{-1mm}
\paragraph{All evaluation dimensions are complementary, rather than interchangeable.}  
As demonstrated in \Cref{fig:radar-chart}, no single metric is fully indicative of all other metrics. This highlights the necessity for researchers and practitioners to select evaluation metrics that are aligned with the specific objectives of their tasks, rather than relying on a single dimension to evaluate the performance of tabular generative models.  
For instance, in regression tasks, BN excels in capturing the underlying tabular data structure, suggesting that its synthetic data can facilitate more accurate causal inference compared to TabDDPM. However, if a practitioner’s primary concern is downstream performance, TabDDPM would be the preferred choice.  
Similarly, SMOTE consistently achieves competitive results in downstream utility across tasks. Nevertheless, SMOTE introduces high risks of privacy leakage, which may be unacceptable for certain sensitive scenarios.

\vspace{-1mm}
\paragraph{Limitations and future work.}
While TabStruct provides valuable insights, we acknowledge several directions for future exploration. 
One primary limitation of TabStruct is the scope of datasets.
As discussed in \Cref{sec:structural-fidelity}, due to the limitations of existing causal discovery methods, TabStruct relies on datasets with expert-validated causal graphs. However, most real-world tabular datasets lack ground truth causal graphs, making it challenging to assess structural fidelity in such cases. 
To address this, we plan to develop new evaluation metrics that enable more flexible assessment of structural fidelity in real-world datasets, where ground truth causal structures are unavailable.


\section{Conclusion}  
\looseness-1
We introduce TabStruct, a novel benchmark framework for the holistic evaluation of tabular generative models.
TabStruct positions structure fidelity as a core aspect of model performance, and quantifies it at the Markov equivalent class level by evaluating the conditional independence relationships between features. Additionally, TabStruct provides conventional evaluation metrics while considering their interplay between structural fidelity. Our experimental results demonstrate that conventional evaluation dimensions fail to provide a holistic view of model performance, and the existing tabular generative models still struggle to effectively capture the underlying structure of tabular data. 
The insights from TabStruct and the open-source library can guide researchers in developing next-generation tabular data generators, and help practitioners select appropriate models for their tasks.


\section*{Acknowledgements}
The authors would like to express their gratitude to Prof. Carl Henrik Ek for insightful discussions on structure learning, and to Prof. Ferenc Huszár and Dr. Ruibo Tu for their enlightening perspectives on causal machine learning.
NS and MJ acknowledge the support of the U.S. Army Medical Research and Development Command of the Department of Defense; through the FY22 Breast Cancer Research Program of the Congressionally Directed Medical Research Programs, Clinical Research Extension Award GRANT13769713. Opinions, interpretations, conclusions, and recommendations are those of the authors and are not necessarily endorsed by the Department of Defense.

\bibliography{iclr2025_delta}
\bibliographystyle{iclr2025_delta}

\newpage
\appendix

\hypersetup{linkcolor=black}
\addcontentsline{toc}{section}{Appendix}
\part{Appendix: \titlecontent}

\mtcsetdepth{parttoc}{3} 
\parttoc
\hypersetup{linkcolor=blue}
\newpage

\section{Summary of Existing Benchmarks}
\label{appendix:literature-review}
\Cref{tab:literature} presents a comparative analysis of TabStruct against existing benchmarks for evaluating tabular generative models. TabStruct is the only benchmark that covers four evaluation metrics, including density estimation, downstream utility, privacy preservation, and structural fidelity. Moreover, it is the only comprehensive benchmark, supporting all eight generator types and offering a more holistic overview of existing tabular generative models. 

\begin{table}[!h]
\centering
\caption{\textbf{Comparison of evaluation scopes between TabStruct and existing benchmarks.} (a)~TabStruct introduces a novel benchmark for the holistic evaluation of tabular generative models, with a particular emphasis on capturing the underlying structure of tabular data. (b)~TabStruct stands out as the only benchmark that covers eight generator categories.
}
\label{tab:literature}
   \subfloat[Evaluation Metrics\label{tab:literature-eval-dim}]{
     \resizebox{\textwidth}{!}{%
\begin{tabular}{l|cc|cc|c|c}
\toprule
\multirow{2}{*}{\textbf{Benchmark source}} & \multicolumn{2}{c|}{\textbf{Density   Estimation}}                             & \multicolumn{2}{c|}{\textbf{Downstream   Utility}}                             & \multirow{2}{*}{\textbf{Privacy   Preservation}} & \multirow{2}{*}{\textbf{Structural Fidelity}} \\
                                                & \textbf{Low-order }                   & \textbf{High-order}                   & \textbf{Classification}               & \textbf{Regression}                   &                                                  &                                                \\
\midrule                                                
\citet{xu2019modeling}                           & {\color[HTML]{3A7B21} \CheckmarkBold} & {\color[HTML]{3A7B21} \CheckmarkBold} & {\color[HTML]{3A7B21} \CheckmarkBold} & {\color[HTML]{3A7B21} \CheckmarkBold} & {\color[HTML]{E6341C} \XSolidBrush}              & {\color[HTML]{E6341C} \XSolidBrush}            \\
\citet{durkan2019neural}                              & {\color[HTML]{3A7B21} \CheckmarkBold} & {\color[HTML]{E6341C} \XSolidBrush}   & {\color[HTML]{E6341C} \XSolidBrush}   & {\color[HTML]{E6341C} \XSolidBrush}   & {\color[HTML]{E6341C} \XSolidBrush}              & {\color[HTML]{E6341C} \XSolidBrush}            \\
\citet{watson2023adversarial}                                    & {\color[HTML]{E6341C} \XSolidBrush}   & {\color[HTML]{E6341C} \XSolidBrush}   & {\color[HTML]{3A7B21} \CheckmarkBold} & {\color[HTML]{E6341C} \XSolidBrush}   & {\color[HTML]{E6341C} \XSolidBrush}              & {\color[HTML]{E6341C} \XSolidBrush}            \\
\citet{liu2023goggle}                                 & {\color[HTML]{3A7B21} \CheckmarkBold} & {\color[HTML]{3A7B21} \CheckmarkBold} & {\color[HTML]{3A7B21} \CheckmarkBold} & {\color[HTML]{E6341C} \XSolidBrush}   & {\color[HTML]{E6341C} \XSolidBrush}              & {\color[HTML]{E6341C} \XSolidBrush}            \\
\citet{borisovlanguage}                                  & {\color[HTML]{3A7B21} \CheckmarkBold} & {\color[HTML]{3A7B21} \CheckmarkBold} & {\color[HTML]{3A7B21} \CheckmarkBold} & {\color[HTML]{3A7B21} \CheckmarkBold} & {\color[HTML]{3A7B21} \CheckmarkBold}            & {\color[HTML]{E6341C} \XSolidBrush}            \\
\citet{kotelnikov2023tabddpm}                                & {\color[HTML]{3A7B21} \CheckmarkBold} & {\color[HTML]{E6341C} \XSolidBrush}   & {\color[HTML]{3A7B21} \CheckmarkBold} & {\color[HTML]{3A7B21} \CheckmarkBold} & {\color[HTML]{3A7B21} \CheckmarkBold}            & {\color[HTML]{E6341C} \XSolidBrush}            \\
\citet{hansen2023reimagining}     & {\color[HTML]{3A7B21} \CheckmarkBold} & {\color[HTML]{3A7B21} \CheckmarkBold} & {\color[HTML]{3A7B21} \CheckmarkBold} & {\color[HTML]{E6341C} \XSolidBrush}   & {\color[HTML]{E6341C} \XSolidBrush}              & {\color[HTML]{E6341C} \XSolidBrush}            \\
\citet{zhangmixed}                                 & {\color[HTML]{3A7B21} \CheckmarkBold} & {\color[HTML]{3A7B21} \CheckmarkBold} & {\color[HTML]{3A7B21} \CheckmarkBold} & {\color[HTML]{3A7B21} \CheckmarkBold} & {\color[HTML]{3A7B21} \CheckmarkBold}            & {\color[HTML]{E6341C} \XSolidBrush}            \\
\citet{tu2024causality}                            & {\color[HTML]{3A7B21} \CheckmarkBold} & {\color[HTML]{3A7B21} \CheckmarkBold} & {\color[HTML]{E6341C} \XSolidBrush}   & {\color[HTML]{E6341C} \XSolidBrush}   & {\color[HTML]{E6341C} \XSolidBrush}              & {\color[HTML]{3A7B21} \CheckmarkBold}          \\
\citet{shi2024tabdiff}                                & {\color[HTML]{3A7B21} \CheckmarkBold} & {\color[HTML]{3A7B21} \CheckmarkBold} & {\color[HTML]{3A7B21} \CheckmarkBold} & {\color[HTML]{3A7B21} \CheckmarkBold} & {\color[HTML]{3A7B21} \CheckmarkBold}            & {\color[HTML]{E6341C} \XSolidBrush}            \\
\midrule
\rowcolor{Gainsboro!60}
\textbf{TabStruct (Ours)}                                   & {\color[HTML]{3A7B21} \CheckmarkBold} & {\color[HTML]{3A7B21} \CheckmarkBold} & {\color[HTML]{3A7B21} \CheckmarkBold} & {\color[HTML]{3A7B21} \CheckmarkBold} & {\color[HTML]{3A7B21} \CheckmarkBold}            & {\color[HTML]{3A7B21} \CheckmarkBold}         \\
\bottomrule
\end{tabular}%
}
   }
   \hfill
   \subfloat[Generator Category Coverage\label{tab:literature-generator}]{
     \resizebox{\textwidth}{!}{%
\begin{tabular}{l|cccccccc|r}
\toprule
Benchmark source                         & Interpolation                         & BN                                    & GAN                                   & VAE                                   & NF                     & Tree                                  & Diffusion                             & LLM                        & \# Generators \\
\midrule
\citet{xu2019modeling}                               & {\color[HTML]{E6341C} \XSolidBrush}   & {\color[HTML]{3A7B21} \CheckmarkBold} & {\color[HTML]{3A7B21} \CheckmarkBold} & {\color[HTML]{3A7B21} \CheckmarkBold} & {\color[HTML]{E6341C} \XSolidBrush}   & {\color[HTML]{E6341C} \XSolidBrush}   & {\color[HTML]{E6341C} \XSolidBrush}   & {\color[HTML]{E6341C} \XSolidBrush}   & 7            \\
\citet{durkan2019neural}                              & {\color[HTML]{E6341C} \XSolidBrush}   & {\color[HTML]{E6341C} \XSolidBrush}   & {\color[HTML]{E6341C} \XSolidBrush}   & {\color[HTML]{3A7B21} \CheckmarkBold} & {\color[HTML]{3A7B21} \CheckmarkBold} & {\color[HTML]{E6341C} \XSolidBrush}   & {\color[HTML]{E6341C} \XSolidBrush}   & {\color[HTML]{E6341C} \XSolidBrush}   & 10           \\
\citet{watson2023adversarial}                                & {\color[HTML]{E6341C} \XSolidBrush}   & {\color[HTML]{E6341C} \XSolidBrush}   & {\color[HTML]{3A7B21} \CheckmarkBold} & {\color[HTML]{3A7B21} \CheckmarkBold} & {\color[HTML]{E6341C} \XSolidBrush}   & {\color[HTML]{3A7B21} \CheckmarkBold} & {\color[HTML]{E6341C} \XSolidBrush}   & {\color[HTML]{E6341C} \XSolidBrush}   & 6            \\
\citet{liu2023goggle}                             & {\color[HTML]{E6341C} \XSolidBrush}   & {\color[HTML]{3A7B21} \CheckmarkBold} & {\color[HTML]{3A7B21} \CheckmarkBold} & {\color[HTML]{3A7B21} \CheckmarkBold} & {\color[HTML]{3A7B21} \CheckmarkBold} & {\color[HTML]{E6341C} \XSolidBrush}   & {\color[HTML]{E6341C} \XSolidBrush}   & {\color[HTML]{E6341C} \XSolidBrush}   & 7            \\
\citet{borisovlanguage}                              & {\color[HTML]{E6341C} \XSolidBrush}   & {\color[HTML]{E6341C} \XSolidBrush}   & {\color[HTML]{3A7B21} \CheckmarkBold} & {\color[HTML]{3A7B21} \CheckmarkBold} & {\color[HTML]{E6341C} \XSolidBrush}   & {\color[HTML]{E6341C} \XSolidBrush}   & {\color[HTML]{E6341C} \XSolidBrush}   & {\color[HTML]{3A7B21} \CheckmarkBold} & 4            \\
\citet{kotelnikov2023tabddpm}                             & {\color[HTML]{3A7B21} \CheckmarkBold} & {\color[HTML]{E6341C} \XSolidBrush}   & {\color[HTML]{3A7B21} \CheckmarkBold} & {\color[HTML]{3A7B21} \CheckmarkBold} & {\color[HTML]{E6341C} \XSolidBrush}   & {\color[HTML]{E6341C} \XSolidBrush}   & {\color[HTML]{3A7B21} \CheckmarkBold} & {\color[HTML]{E6341C} \XSolidBrush}   & 6            \\
\citet{hansen2023reimagining} & {\color[HTML]{E6341C} \XSolidBrush}   & {\color[HTML]{3A7B21} \CheckmarkBold} & {\color[HTML]{3A7B21} \CheckmarkBold} & {\color[HTML]{3A7B21} \CheckmarkBold} & {\color[HTML]{3A7B21} \CheckmarkBold} & {\color[HTML]{E6341C} \XSolidBrush}   & {\color[HTML]{3A7B21} \CheckmarkBold} & {\color[HTML]{E6341C} \XSolidBrush}   & 5            \\
\citet{zhangmixed}                             & {\color[HTML]{3A7B21} \CheckmarkBold} & {\color[HTML]{E6341C} \XSolidBrush}   & {\color[HTML]{3A7B21} \CheckmarkBold} & {\color[HTML]{3A7B21} \CheckmarkBold} & {\color[HTML]{E6341C} \XSolidBrush}   & {\color[HTML]{E6341C} \XSolidBrush}   & {\color[HTML]{3A7B21} \CheckmarkBold} & {\color[HTML]{3A7B21} \CheckmarkBold} & 9            \\
\citet{tu2024causality}                       & {\color[HTML]{E6341C} \XSolidBrush}   & {\color[HTML]{E6341C} \XSolidBrush}   & {\color[HTML]{3A7B21} \CheckmarkBold} & {\color[HTML]{3A7B21} \CheckmarkBold} & {\color[HTML]{E6341C} \XSolidBrush}   & {\color[HTML]{E6341C} \XSolidBrush}   & {\color[HTML]{3A7B21} \CheckmarkBold} & {\color[HTML]{3A7B21} \CheckmarkBold} & 7            \\
\citet{shi2024tabdiff}                            & {\color[HTML]{E6341C} \XSolidBrush}   & {\color[HTML]{E6341C} \XSolidBrush}   & {\color[HTML]{3A7B21} \CheckmarkBold} & {\color[HTML]{3A7B21} \CheckmarkBold} & {\color[HTML]{E6341C} \XSolidBrush}   & {\color[HTML]{E6341C} \XSolidBrush}   & {\color[HTML]{3A7B21} \CheckmarkBold} & {\color[HTML]{3A7B21} \CheckmarkBold} & 9            \\
\midrule
\rowcolor{Gainsboro!60}
\textbf{TabStruct (Ours)}                               & {\color[HTML]{3A7B21} \CheckmarkBold} & {\color[HTML]{3A7B21} \CheckmarkBold} & {\color[HTML]{3A7B21} \CheckmarkBold} & {\color[HTML]{3A7B21} \CheckmarkBold} & {\color[HTML]{3A7B21} \CheckmarkBold} & {\color[HTML]{3A7B21} \CheckmarkBold} & {\color[HTML]{3A7B21} \CheckmarkBold} & {\color[HTML]{3A7B21} \CheckmarkBold} & \textbf{9}    \\
\bottomrule
\end{tabular}%
}
   }
\end{table}

\newpage
\section{Reproducibility}
\label{appendix:reproducibility}

\subsection{Reference Datasets}
\label{appendix:dataset}
\looseness-1
To ensure that the causal structures of reference datasets are realistic, we select seven publicly available datasets from bnlearn~\citep{scutari2011bnlearn}. Each dataset is accompanied by a ground-truth structural causal model (SCM) validated by human experts. Furthermore, to obtain generalisable benchmark results, we select datasets from diverse domains, and they are across three different levels of structure scales (i.e., small, medium and large). 

In contrast, the only prior benchmark that addresses structural fidelity, CauTabBench~\citep{tu2024causality}, does not utilise SCMs validated by human experts. Additionally, the dimensionality of their datasets is fixed at 10 numerical features. In summary, TabStruct is one of the first to offer a comprehensive benchmark for tabular generative models, leveraging datasets with realistic causal structures, mixed feature types, and more than 10 features.

\begin{table}[!htbp]
\setlength{\tabcolsep}{2pt}
\centering
\caption{Details of four classification datasets with realistic structures.}
\label{tab:dataset-classification-structure}
\resizebox{\textwidth}{!}{%
\begin{tabular}{lllrrrrrrr}
\toprule
Dataset    & Domain      & Structure scale                       & \# Samples & \# Features & \# Numerical & \# Categorical & \# Classes & \begin{tabular}[c]{@{}r@{}}\# Samples per class \\      (Min)\end{tabular} & \begin{tabular}[c]{@{}r@{}}\# Samples per class \\      (Max)\end{tabular} \\
\midrule
Sangiovese & Agriculture & Small (\textless{}20 nodes) & 2,000      & 15          & 14           & 1              & 16         & 108                                                                        & 146                                                                        \\
Insurance  & Economics   & Medium (20–50 nodes)        & 2,000      & 27          & 0            & 27             & 4          & 38                                                                         & 1,122                                                                      \\
Hailfinder & Meteorology & Large ($>$50 nodes)        & 2,000      & 56          & 0            & 56             & 3          & 519                                                                        & 880                                                                        \\
ANDES      & Education   & Large ($>$50 nodes) & 2,000      & 223         & 0            & 223            & 2          & 830                                                                        & 1,170    \\
\bottomrule
\end{tabular}%
}
\end{table}

\begin{table}[!htbp]
\centering
\caption{Details of three regression datasets with realistic structures.}
\label{tab:dataset-regression-structure}
\resizebox{\textwidth}{!}{%
\begin{tabular}{lllrrrr}
\toprule
Dataset    & Domain       & Structure scale             & \# Samples & \# Features & \# Numerical & \# Categorical \\
\midrule
Healthcare & Medicine     & Small (\textless{}20 nodes) & 2,000      & 7           & 7            & 3              \\
MEHRA      & Meteorology  & Medium (20–50 nodes)        & 2,000      & 24          & 20           & 4    \\
ARTH150    & Life Science & Large ($>$50 nodes) & 2,000      & 107         & 107          & 0              \\
\bottomrule
\end{tabular}%
}
\end{table}

\subsection{Data Splitting}
\label{appendix:data-splitting}

\begin{figure}[!h]
    \centering
    \includegraphics[width=0.7\textwidth]{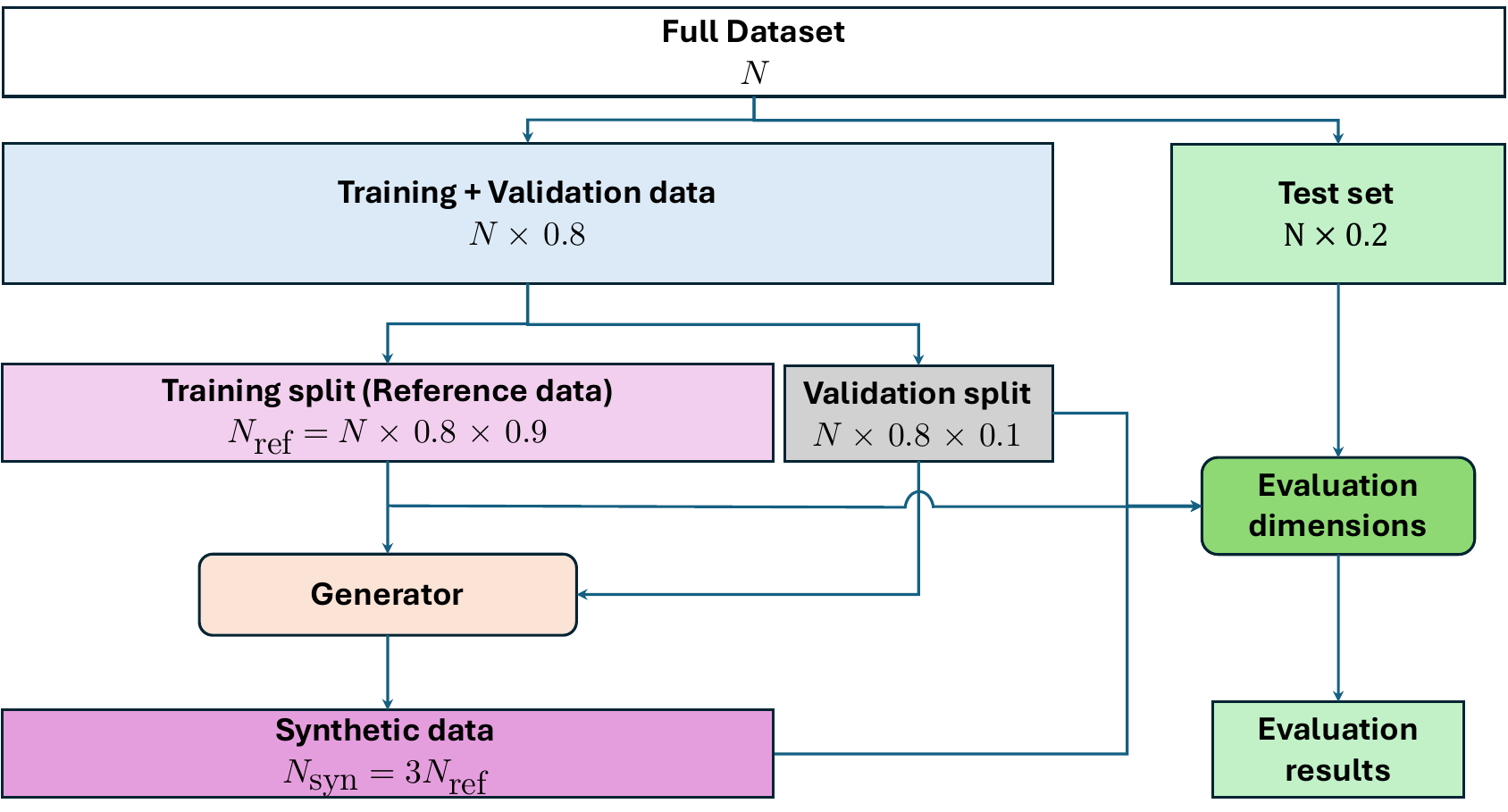}
    \caption{Data splitting strategies for benchmarking tabular data generators.}
\label{fig:splitting}
\end{figure}

\subsection{Data Preprocessing}
\label{appendix:data-preprocessing}

Following the procedures presented in prior work~\citep{mcelfresh2024neural, grinsztajn2022tree}, we perform preprocessing in two steps. Firstly, we impute the missing values with the mean value for numerical features and the most mode value for categorical features. We then compute the required statistics with training data and then transform it. For categorical features, we convert them into one-hot encodings. For numerical features, we perform Z-score normalisation. We compute each feature's mean and standard deviation in the training data and then transform the training samples to have a mean of zero and a variance of one for each feature. Finally, we apply the same transformation to the validation and test data before conducting evaluations.

\subsection{Implementation of conditional independence test}
\label{appendix:implemetation-ci-test}
\looseness-1
For categorical datasets, we employ the chi-square independence test~\citep{mchugh2013chi}; for numerical datasets, we use partial correlation based on the Pearson correlation coefficient~\citep{baba2004partial}; and for mixed datasets, we utilise a residualisation-based conditional independence test~\citep{ankan2023simple, li2010test, muller1984practical}. 
We implement these conditional independence tests using pgmpy~\citep{ankan2024pgmpy}, an open-source Python library for causal and probabilistic inference. The significance level is set to 0.01 by default (i.e., the p-value is 0.01).

\subsection{Implementations of Benchmark Generators}
\label{appendix:implementation-generator}

\textbf{SMOTE} is an interpolation-based oversampling method~\citep{chawla2002smote}. It generates new samples by interpolating between real samples. 
We use the open-source implementation of SMOTE from Imbalanced-learn~\citep{imbalanced-learn}, setting the number of neighbours $k$ within the range $\{1, 3, 5\}$. When applicable, we use the default value for nearest neighbours (i.e., $k=5$).

\textbf{Bayesian Network (BN)} is a probabilistic graphical model used to represent and reason about the dependence relationships between features~\citep{qian2024synthcity, hansen2023reimagining}. It consists of two main components: (i)~a causal discovery model to construct a directed acyclic graph (DAG), where features and the target serve as nodes, and their dependencies are represented as edges; (ii)~a parameter estimation mechanism to quantify the dependence relationships. Following \citet{hansen2023reimagining}, the causal discovery method is selected from Hill Climbing Search~\citep{koller2009probabilistic}, the Peter-Clark algorithm~\citep{koller2009probabilistic}, and Chow-Liu or Tree-augmented Naive Bayes~\citep{chow1968approximating, friedman1997bayesian}. We then build the parametrised BN using maximum likelihood estimation.

\textbf{TVAE} is a variational autoencoder (VAE) designed for tabular data~\citep{xu2019modeling}. TVAE employs mode-specific normalisation to handle the complex distributions of numerical features. To address the class imbalance problem, TVAE conditions on specific categorical features during generation.

\textbf{GOGGLE} is a VAE-based tabular data generator designed to model the dependence relationships between features~\citep{liu2023goggle}. GOGGLE proposes to learn an adjacency matrix to model the dependence relationships between features. However, TabStruct and prior benchmarks~\citep{margeloiutabebm, zhangmixed, shi2024tabdiff} all show that the downstream utility of GOGGLE is limited. We hypothesise that this is because of the challenge of learning accurate structures of tabular data. The inherent structure learning mechanism in GOGGLE fails to capture accurate conditional independence relationships between features, it could thus lead to poor-quality synthetic data, even if the model attempts to explicitly model the relationships between features like GOGGLE.

\textbf{CTGAN} is a conditional generative adversarial network (GAN) designed for tabular data~\citep{xu2019modeling}. CTGAN leverages PacGAN~\citep{lin2018pacgan} framework to mitigate mode collapse. In addition, CTGAN employs the same mode-specific normalisation technique as TVAE.

\textbf{NFlow} is a normalisation flow model designed for tabular data generation~\citep{durkan2019neural}. NFlow incorporates neural splines as a drop-in replacement for affine or additive transformations in coupling and autoregressive layers.

\textbf{TabDDPM} is a diffusion-based model for tabular data generation~\citep{kotelnikov2023tabddpm}. TabDDPM introduces two core diffusion processes: (i)~Gaussian noise for numerical features and (ii)~multinomial diffusion with categorical noise for categorical features. TabDDPM directly concatenates numerical and categorical features as the input and output of the denoising function.

\textbf{ARF} is a tree-based model for tabular data generation~\citep{watson2023adversarial}. ARF employs a recursive adaptation of unsupervised random forests for joint density estimation by iteratively refining synthetic data distributions using adversarial training principles.

\textbf{GReaT} leverages large language models (LLMs) to generate synthetic tabular data~\citep{borisovlanguage}. GReaT converts each sample into a sentence and fine-tunes the language model to capture the sentence-level distributions. Additionally, GReaT shuffles the order of features to mitigate the permutation variance in sentence-level distributions.

\subsection{Software and Computing Resources}
\label{appendix:software-and-computing}

\looseness-1
\textbf{Software implementation.} 
\textit{(i)~For generators:}  We implemented SMOTE with Imbalanced-learn~\citep{imbalanced-learn}, an open-source Python library for imbalanced datasets with an MIT licence. For other benchmark generators, we used their open-source implementations in Synthcity~\citep{qian2024synthcity}, a library for generating and evaluating synthetic tabular data with an Apache-2.0 license. 
\textit{(ii)~For downstream predictors:} We implemented TabPFN with its open-source implementation (\url{https://github.com/automl/TabPFN}). We implemented the other five downstream predictors (i.e., Logistic Regression, KNN, MLP, Random Forest and XGBoost) with their open-source implementation in scikit-learn~\citep{scikit-learn}, an open-source Python library under the 3-Clause BSD license. 
\textit{(iii)~For result analysis and visualisation:} All numerical plots and graphics have been generated using Matplotlib 3.7~\citep{matplotlib}, a Python-based plotting library with a BSD licence. The icons for downstream tasks are from \url{https://icons8.com/}.

We ensure the consistency and reproducibility of experimental results by implementing a uniform pipeline using PyTorch Lightning, an open-source library under an Apache-2.0 licence. We further fixed the random seeds for data loading and evaluation throughout the training and evaluation process. This ensured that TabEBM and all benchmark models were trained and evaluated on the same set of samples. The experimental environment settings, including library dependencies, are specified in the open-source library for reference and reproduction purposes.

\textbf{Computing Resources.}
All the experiments were conducted on a machine equipped with an NVIDIA A100 GPU with 40GB memory and an Intel(R) Xeon(R) CPU (at 2.20GHz) with six cores. The operating system used was Ubuntu 20.04.5 LTS.

\newpage
\section{Rationales for Sample Size of Synthetic Data}
\label{appendix:syn-sample-size}

\begin{figure}[!htbp]
    \centering
    \includegraphics[width=0.7\textwidth]{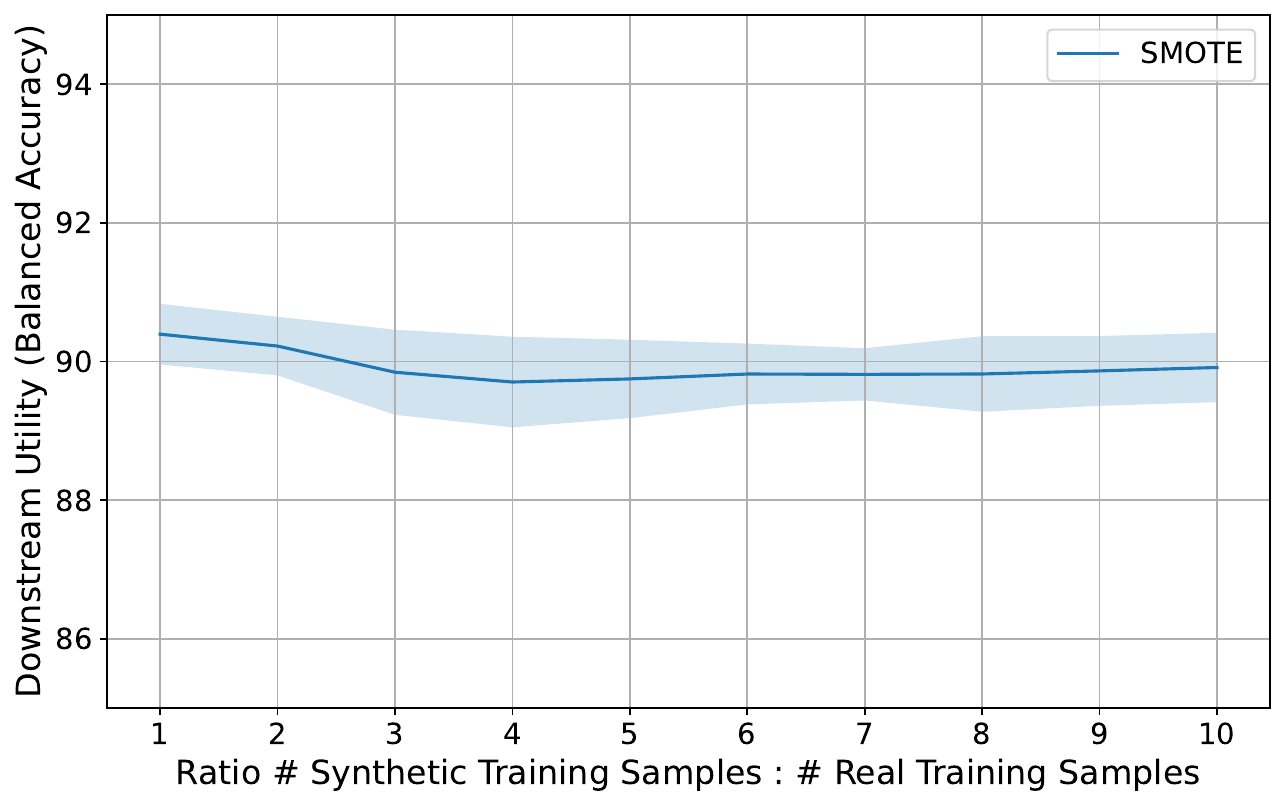}
    \caption{\textbf{Downstream utility vs. different ratios between the number of synthetic data and reference data ($N_{\text{syn}}:N_{\text{ref}}$).} On the ``Hailfinder'' dataset, as $N_{\text{syn}}$ increases, the evaluation results become saturated. Specifically, the range of balanced accuracy varies by less than 0.3\% when the ratio increases from $N_{\text{syn}}:N_{\text{ref}}=3:1$ to $N_{\text{syn}}:N_{\text{ref}}=10:1$. Therefore, we set $N_{\text{syn}}=3N_{\text{ref}}$ in all experiments to ensure stable evaluation results.}
\label{fig:saturation}
\end{figure}

\end{document}